\documentclass{article}
\usepackage{amsmath,amsfonts,bm}

\def\eqref#1{equation~\ref{#1}}
\def\1{\bm{1}}

\def\rve{{\mathbf{e}}}

\def\rvp{{\mathbf{p}}}

\def\rvt{{\mathbf{t}}}

\def\rvx{{\mathbf{x}}}
\def\rvy{{\mathbf{y}}}

\DeclareMathAlphabet{\mathsfit}{\encodingdefault}{\sfdefault}{m}{sl}
\SetMathAlphabet{\mathsfit}{bold}{\encodingdefault}{\sfdefault}{bx}{n}

\def\gA{{\mathcal{A}}}

\def\gH{{\mathcal{H}}}

\def\gS{{\mathcal{S}}}

\newcommand{\E}{\mathbb{E}}

\DeclareMathOperator*{\expect}{\mathbb{E}}

\usepackage{graphicx}
\usepackage{amsmath}
\usepackage{amssymb}
\usepackage{booktabs}
\usepackage{algorithmic}
\usepackage{cite}
\usepackage[table]{xcolor}
\usepackage{colortbl}
\usepackage{textcomp}
\usepackage{wrapfig}
\usepackage{xcolor}
\usepackage{multicol}
\usepackage{multirow}
\usepackage{array}
\usepackage{tabularx}
\usepackage{rotating}
\usepackage{xcolor}
\usepackage{hyperref}
\usepackage[bottom]{footmisc}
\usepackage{caption}
\usepackage{array}
\usepackage[ruled,linesnumbered]{algorithm2e}
\usepackage{mwe}
\usepackage[justification=centering]{subfig}
\usepackage[shortlabels]{enumitem}
\usepackage[normalem]{ulem}
\usepackage{float}
\usepackage{spconf,amsmath,graphicx}
\usepackage{url}
\usepackage{enumitem}
\setlist{nosep, leftmargin=14pt}
\usepackage[T1]{fontenc}
\usepackage{times}
\usepackage{mwe} % to get dummy images
\title{Can language-guided unsupervised adaptation improve medical image classification using unpaired images and texts? }
\name{Umaima Rahman$^{1}$, Raza Imam$^{1}$, Mohammad Yaqub$^{1}$, Boulbaba Ben Amor$^{2}$, Dwarikanath Mahapatra$^{3}$}
\address{$^{1}$Mohamed Bin Zayed University of Artificial Intelligence, $^{2}$Inception, UAE \\ 
$^{3}$Monash University, Australia.}
\begin{document}
%\ninept
%
\maketitle

\begin{abstract}
% In medical image classification, supervised learning is challenging due to the lack of abundant labeled medical images. To overcome this, we can leverage the visual-textual alignment within VLMs (VLMs) to aid the unsupervised learning of a medical image classifier. In this work, we propose \underline{Med}ical \underline{Un}supervised \underline{A}daptation (\texttt{MedUnA}) of VLMs, where the LLM-generated descriptions corresponding to a class label are passed through a text encoder, and the resulting text embeddings are matched with the class label by training a cross-modal adapter. The adapter is then attached to a visual encoder of \texttt{MedCLIP} to align the visual embeddings through an unsupervised learning scheme driven by a contrastive entropy-based loss and prompt tuning. It is also worth noting that unlike traditional VLMs, \texttt{MedUnA} uses unpaired image and text to learn vision and language representations. We evaluate the performance of \texttt{MedUnA} on three chest X-ray datasets and two multi-class datasets of diabetic retinopathy and skin lesions. The results demonstrate significant average accuracy gains across all five medical datasets compared to the Zero-Shot baseline, highlighting the efficacy of our approach. The code shall be made publicly available upon acceptance.
In medical image classification, supervised learning is challenging due to the scarcity of labeled medical images. To address this, we leverage the visual-textual alignment within Vision-Language Models (VLMs) to enable unsupervised learning of a medical image classifier. In this work, we propose \underline{Med}ical \underline{Un}supervised \underline{A}daptation (\texttt{MedUnA}) of VLMs, where the LLM-generated descriptions for each class are encoded into text embeddings and matched with class labels via a cross-modal adapter. This adapter attaches to a visual encoder of \texttt{MedCLIP} and aligns the visual embeddings through unsupervised learning, driven by a contrastive entropy-based loss and prompt tuning. Thereby, improving performance in scenarios where textual information is more abundant than labeled images, particularly in the healthcare domain. Unlike traditional VLMs, \texttt{MedUnA} uses \textbf{unpaired images and text} for learning representations and enhances the potential of VLMs beyond traditional constraints. We evaluate the performance on three chest X-ray datasets and two multi-class datasets (diabetic retinopathy and skin lesions), showing significant accuracy gains over the zero-shot baseline. Our code is available at https://github.com/rumaima/meduna.

\end{abstract}
\vspace{-2mm}
\section{Introduction}
\label{sec:introduction}
\vspace{-2mm}
A major challenge in the field of medical imaging is the scarcity of high-quality labeled medical data due to the need for expert annotations. This limits the performance of machine learning models for medical image tasks that rely on large amounts of labeled data for training. Labeled visual data such as CT scans, X-rays, and MRIs often require access permissions and are subject to privacy regulations. On the contrary,
there is a greater abundance of disease-related information in the form of textual data, which is easier to access than labeled medical images. This includes data mined from anonymized electronic health records, medical reports, and transcriptions. As a result, instead of using an image-based training method, we can leverage the visual-textual alignment within VLMs to facilitate an unsupervised learning regime.

% general knowledge in the form of textual data for medical diagnosis is more readily available than labeled images and accessible through data mining from anonymized electronic health records, medical reports, and transcriptions. 

\vspace{0.15cm}
\noindent\textbf{Motivation:}
During training a VLM, the textual and visual embeddings related to a sample become typically close in the embedding space due to the contrastive training objective. The shift between the semantic and visual embeddings is defined as the \textit{modality gap} \cite{zhang2022drml}. Consequently, a low modality gap would imply that a classifier trained on textual embeddings can be used for inference on visual embeddings. Minimizing the modality gap has the potential to enhance the learning process and improving the model performance by enabling more effective integration of multi-modal data.

% With the aim of minimizing this modality gap, we can initialize an adapter through supervised training on text embeddings of different class labels and then tune it through unsupervised learning using visual embeddings of unlabeled samples.

\vspace{-0.5mm}
\noindent\textbf{Related Works:}
The current body of research on VLMs for medical images demonstrates a prevalent trend where the emphasis is on the pretrain-finetune paradigm. This involves pre-training a VLM on large datasets constituting of image-text pairs, followed by fine-tuning for specific downstream tasks. For example, \cite{huix2024natural} investigates transferring pre-trained foundation models to medical image classification tasks using two fine-tuning strategies. The authors in \cite{zhan2024unidcp} propose a unified model for multiple vision-language tasks but relies on extensive pre-training, whereas \cite{qin2024freeze} introduces a backbone-agnostic adapter framework that combines frozen pre-trained encoders with lightweight adapters to facilitate cross-modal learning. However, all these approaches assume the availability of image-text pairs, which limits the applicability of VLMs when unpaired data is solely available. Transitioning to unsupervised data is essential for several reasons: Firstly, the use of unpaired medical images without the corresponding textual annotations makes the process scalable. Secondly, unsupervised models trained on unpaired medical data generalize well across diverse clinical scenarios, especially when obtaining labeled datasets can be challenging. Thirdly, using unpaired data reduces the time and resources required for manual annotation, facilitating timely diagnosis. 

\begin{figure*}[t!]
    \centering
    \includegraphics[height=0.40\textwidth]{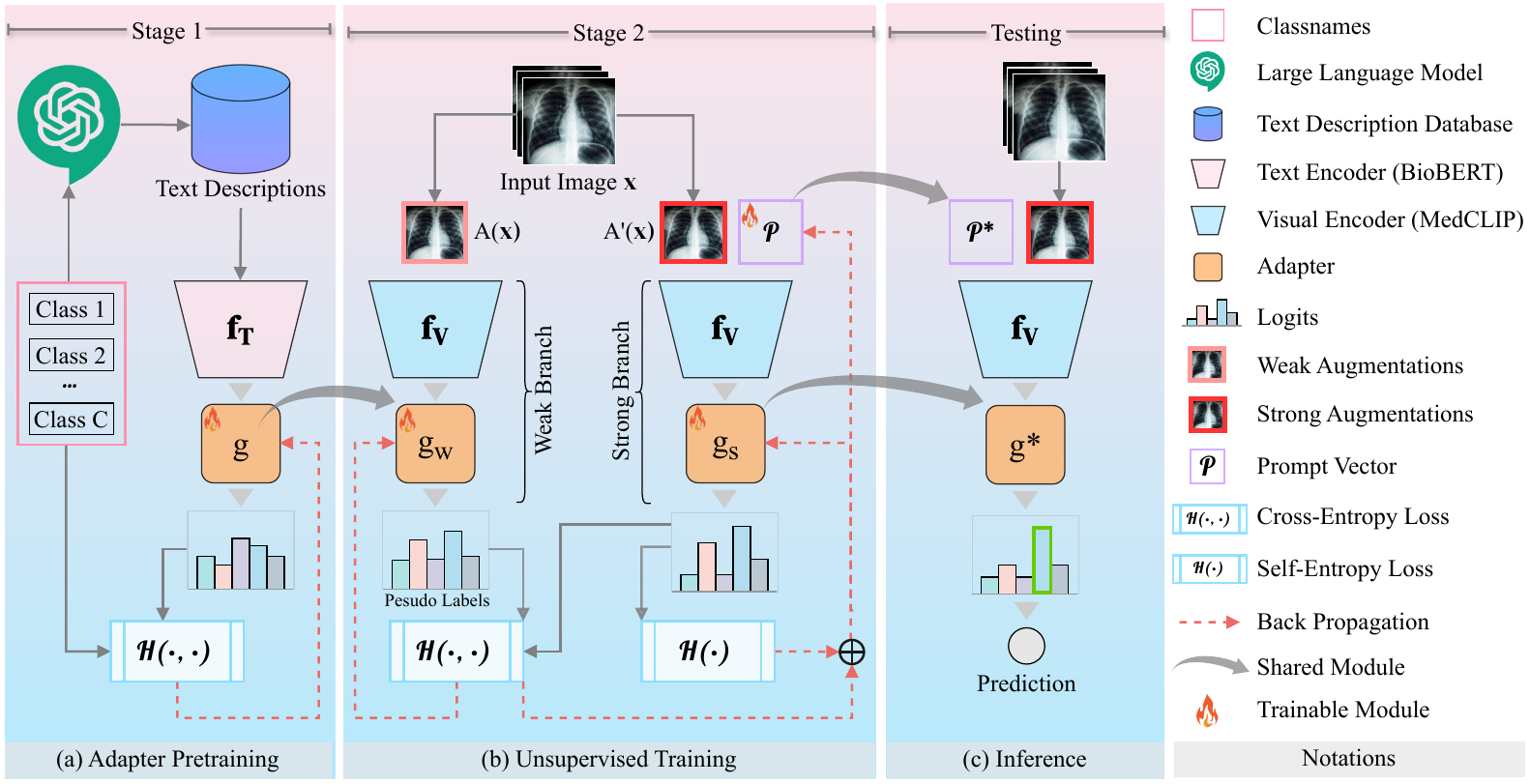}
    \vspace{-2.5mm}
    \caption{\small \textbf{The \texttt{MedUnA} framework:} (a) Adapter Pre-training: A textual classifier is trained to classify the LLM-generated descriptions for a disease. (b) Unsupervised Training: The trained textual classifier \& a learnable prompt vector for an unpaired image embedding is trained in an unsupervised regime. (c) Inference: The tuned textual classifier \& the prompt vector are used to get predictions on the test dataset.}
    \label{fig:MedUnA_workflow}
\end{figure*}

\vspace{-0.5mm}
\noindent\textbf{Contribution:} (1) Motivated by the abundance of medical textual data, we propose to use a label-free tuning (LFT) based unsupervised approach as illustrated in Fig. \ref{fig:MedUnA_workflow}. (2) In order to alleviate the pre-training requirements on vast medical datasets of image-text pairs, we propose a language-guided, annotation-free unsupervised adaptation of VLMs for the task of medical image classification. (3)  By avoiding extensive pre-training of VLMs on large medical datasets, \texttt{MedUnA} makes effective use of the parameter space in the memory along with an increase in accuracy. This is done by training a lightweight adapter on a textual corpus created using descriptions of novel disease classes as the first step. In the second step, this adapter and a learnable prompt vector are further trained using the visual embeddings for the unlabeled images using our entropy minimization-based loss. 
\vspace{-3mm}
\section{Methodology}
\label{sec:methodology}
\vspace{-3mm}
Consider a set of medical images $\rvx \in \gS$ which are unlabeled, where $\gS$ is the dataset. Also, consider a finite set of classes $[C]$ to which each image in $\gS$ can be assigned.
\vspace{-4mm}
\subsection{Adapter Pre-training}
In \textbf{Stage 1}, we first generate textual descriptions associated with class $k \in [C]$ by designing suitable prompt $\rvt_k = \mathfrak{P}(l_k)$, where $l_k$ is the label associated with class $k \in [C]$, respectively. The prompts consider critical information such as disease nomenclature, etiology, patho-physiology, clinical manifestations, and diagnostic criteria.
For a certain class $k$, an LLM is used to generate the textual description 
$\mathfrak{L}(l_k, \rvt_k)$, which is then embedded by the text encoder $f_T(\cdot)$ as $f_T \circ \mathfrak{L}(l_k, \rvt_k)$. This embedding is then passed through the cross-modal adapter $g(\cdot)$ to generate the class logits associated with the input label $l_k$ and can be written as $g \circ f_T \circ \mathfrak{L}(l_k, \rvt_k)$. 
The cross-modal adapter is trained by minimizing the cross-entropy between $g \circ f_T \circ \mathfrak{L}(l_k, \rvt_k)$, and the ground truth $\rve_k$, which denotes the $k^{\rm th}$ standard basis vector of $\mathbb{R}^C$.
Subsequently we obtain the cross-modal adapter $\hat{g}$ through:
\begin{align}
  \min_{g} \mathbb{E}_{k \in [C]} \gH( \mathbf{e}_k, g \circ f_T \circ \mathfrak{L} (l_k)), 
\end{align}
where $\gH(\rvy, \rvy') \triangleq -\rvy^\top \log(\rvy')$ denotes cross-entropy.

{\renewcommand{\arraystretch}{1.0}
\begin{table*}[t]
\centering
\resizebox{0.90\textwidth}{!}{%E
\begin{tabular}{l|cccc|cccc}
\toprule
Backbone $\rightarrow$ & \multicolumn{4}{c}{\texttt{CLIP-ViT-B/32}} & \multicolumn{4}{|c}{\texttt{MedCLIP-Swin}} \\ \cmidrule(r){1-1} \cmidrule(lr){2-5} \cmidrule(l){6-9} 
Dataset $\downarrow$ & \cellcolor[HTML]{F4F8F8}Zero-Shot & \cellcolor[HTML]{EDF7F8}\texttt{LaFTer}\cite{mirza2024lafter} &  \cellcolor[HTML]{E4F8FC}\texttt{TPT}\cite{shu2022test} & \cellcolor[HTML]{DBF5FA}\texttt{MedUnA} (Ours) & \cellcolor[HTML]{F4F8F8}Zero-shot & \cellcolor[HTML]{EDF7F8}\texttt{LaFTer}\cite{mirza2024lafter} & \cellcolor[HTML]{E4F8FC}\texttt{TPT}\cite{shu2022test} & \cellcolor[HTML]{DBF5FA}\texttt{MedUnA} (Ours) \\ \midrule

\texttt{S-TB} \cite{shenzhenTB, jaeger2014two}         &\cellcolor[HTML]{F4F8F8}42.86  & \cellcolor[HTML]{EDF7F8}57.14  & \cellcolor[HTML]{E4F8FC}50.76    & \cellcolor[HTML]{DBF5FA}\underline{57.14} &  \cellcolor[HTML]{F4F8F8}42.11           & \cellcolor[HTML]{EDF7F8}42.86 & \cellcolor[HTML]{E4F8FC}54.08 &  \cellcolor[HTML]{DBF5FA}\textbf{67.67}\\

\texttt{M-TB} \cite{montgomeryTB, jaeger2014two}   &\cellcolor[HTML]{F4F8F8}48.28  & \cellcolor[HTML]{EDF7F8}51.72  & \cellcolor[HTML]{E4F8FC}42.03    & \cellcolor[HTML]{DBF5FA}\underline{62.07} &  \cellcolor[HTML]{F4F8F8}55.17           & \cellcolor[HTML]{EDF7F8}48.28 & \cellcolor[HTML]{E4F8FC}57.97 &  \cellcolor[HTML]{DBF5FA}\textbf{72.41}\\

\texttt{G-Pneumonia} \cite{guangzhou} &\cellcolor[HTML]{F4F8F8}62.50  &\cellcolor[HTML]{EDF7F8}62.50  & \cellcolor[HTML]{E4F8FC}\underline{72.97}    & \cellcolor[HTML]{DBF5FA}65.71  & \cellcolor[HTML]{F4F8F8}\textbf{80.93}  & \cellcolor[HTML]{EDF7F8}78.53 & \cellcolor[HTML]{E4F8FC}72.98 &  \cellcolor[HTML]{DBF5FA}76.12 \\

\texttt{IDRID} \cite{idrid, porwal2018indian}  &\cellcolor[HTML]{F4F8F8}20.39  & \cellcolor[HTML]{EDF7F8}32.01  & \cellcolor[HTML]{E4F8FC}14.98    & \cellcolor[HTML]{DBF5FA}\underline{32.01} & \cellcolor[HTML]{F4F8F8} 27.18           & \cellcolor[HTML]{EDF7F8}18.45 & \cellcolor[HTML]{E4F8FC}23.48 &  \cellcolor[HTML]{DBF5FA}\textbf{32.04} \\

\texttt{ISIC} \cite{isicArchive, codella2018skin} &\cellcolor[HTML]{F4F8F8}12.90 & \cellcolor[HTML]{EDF7F8}12.43 & \cellcolor[HTML]{E4F8FC}24.60   & \cellcolor[HTML]{DBF5FA}\underline{25.46} &  \cellcolor[HTML]{F4F8F8}08.80  & \cellcolor[HTML]{EDF7F8}14.42  & \cellcolor[HTML]{E4F8FC}08.69  & \cellcolor[HTML]{DBF5FA}\textbf{29.70}\\ 
\midrule
Average & \cellcolor[HTML]{F4F8F8}47.16 & \cellcolor[HTML]{EDF7F8}43.16 & \cellcolor[HTML]{E4F8FC}41.07 & \cellcolor[HTML]{DBF5FA}\underline{48.48} & \cellcolor[HTML]{F4F8F8}42.84 & \cellcolor[HTML]{EDF7F8}40.51 & \cellcolor[HTML]{E4F8FC}43.44 & \cellcolor[HTML]{DBF5FA}\textbf{55.59} \\
\bottomrule
\end{tabular}}
\vspace{-2.5mm}
\caption{\small Top-1 Test Accuracy (\%) of models evaluated across different datasets using two distinct visual encoders: \texttt{CLIP} and \texttt{MedCLIP}. Our proposed \texttt{MedUnA} framework consistently outperforms other models in overall performance. In the table, the highest accuracy achieved using the \texttt{CLIP} encoder is \underline{underlined}, while the best results obtained with the \texttt{MedCLIP} encoder are highlighted in \textbf{bold}.}
\vspace{-2.5mm}
\label{tab:MedUnA-acc}
\end{table*}}

\subsection{Unsupervised Training} 
Next, in \textbf{Stage 2}, we use the medical images to further train the cross-modal adapter initialized to $\hat{g}$ through an unsupervised approach. We start with two identical branches consisting of the visual encoder followed by adapters initialized with the weights $\hat{g}$ \cite{mirza2024lafter}. In one branch, we apply subtle transformations, such as small rotations, to create weak augmentations $\gA(\rvx)$ of the input $\rvx$,and more aggressive transformations like larger rotations, cropping and intensity scaling to generate strong augmented version $\gA'(\rvx)$ of the input, appended by a learnable prompt vector $\rvp$ in the other branch. Henceforth, we refer to these branches as the \textit{weak} and \textit{strong} branches, respectively. The weak branch generates pseudo labels based on the visual embeddings, while the strong branch computes logits that are aligned with these pseudo labels using a cross-entropy loss.
The pseudo-label associated with image $\rvx$ generated by the weak branch can be written as ${g}_w \circ f_V \circ \gA(\rvx)$, and the corresponding output from the strong branch is:
${g}_s \circ f_V \circ \gA'(\rvx \parallel \rvp )$, where $\rvp$ denotes the learnable prompt, and $f_V(\cdot)$ is the visual encoder.
The weak branch adapter is trained by minimizing the cross-entropy between the outputs of the two branches:
    \begin{align}
    \min_{{g}_w} \E_{\rvx \in \gS} \gH\big( {g}_w \circ f_V \circ \gA(\rvx), {g}_s \circ f_V  (\gA'(\rvx) \parallel \rvp ) \big),
    \end{align}
    where $g_w$ is initialized to $\hat{g}$.
The strong branch adapter $g_s$, and the prompt $\rvp$ are jointly learnt through:
\begin{align}
    \min_{{g}_s, \rvp} \E_{\rvx \in \gS} \gH\big( {g}_w \circ f_V \circ \gA(\rvx), {g}_s \circ f_V  (\gA'(\rvx) \parallel \rvp ) \big) \nonumber \\
    + \lambda \gH\left({g}_s \circ f_V  (\gA'(\rvx) \parallel \rvp )\right), \quad \lambda \in \mathbb{R}^+,
\end{align}
where $g_s$ is initialized to $\hat{g}$.
The additional self-entropy term in the strong branch encourages diversity and balances it with the consistency enforced by cross-entropy term. The motivation behind minimizing entropy \cite{grandvalet2004semi} stems from the intuition that models are generally more accurate when they make high-confidence predictions as shown in \cite{pressentropy}, \cite{wang2020tent}. 
This unsupervised training enhances the model's generalization capabilities and facilitates predictions by utilizing a cross-modal adapter that is trained on unpaired textual descriptions and further tuned with visual information from unlabelled images, making it useful in scenarios where textual data is more abundant than labeled images, such as in the healthcare domain.

% Additionally, we utilize two visual encoders, CLIP (ViT-B/32) and MedCLIP (Swin), where the hidden sizes and the number of tunable parameters vary, allowing our approach to remain parameter-efficient compared to traditional pre-training and fine-tuning strategies that often require tuning a larger number of parameters.

\vspace{-3mm}
\subsection{Inference} 

The resultant strong branch from Stage 2 with adapter $g^\star$, and prompt $\rvp^\star$ is used for inference, where the classification output for input $\rvx$ is:
\vspace{-0.2cm}
\begin{align}
    \arg \max_{k \in [C]} g^{\star} \circ f_V  (\gA'(\rvx) \parallel \rvp^{\star} ).
    \vspace{-0.4cm}
\end{align}
% The resulting \textit{strong} branch with adapter $g^{\star}$ and prompt $\rvp^{\star}$ is then used during \textbf{inference}, where the classification output is $$$$
In the presence of true labels, in a supervised approach \cite{zhang2022drml}, we could jointly train the cross-modal adapter, and the prompt by minimizing the misalignment in the textual and visual embeddings associated with input $\rvx$.
\begin{align}
    \min_{g, \rvp}  \expect_{\rvx \in \gS} \Delta\Big( g \circ \underbrace{f_V( \gA'(\rvx) || \rvp)}_{\text{visual embedding}}, \quad g \circ \underbrace{f_T( \mathfrak{L}(l_{\rvx}) )}_{\text{text embedding}}  \Big),
\end{align}
where $l_{\rvx}$ is the true class label of the image sample $\rvx$, and $\Delta(\cdot, \cdot)$ is a distance measure that quantifies the misalignment in the textual and visual embeddings. However, in the absence of the true labels we resort to the unsupervised method outlined above. This approach enables effective classification even in scenarios where labeled data is scarce or unavailable.
\vspace{-6mm}
\section{Experiments and Results}
\vspace{-2mm}
\subsection{Experimentation Details}
\vspace{-2mm}
We conduct extensive experiments on five publicly available medical datasets to answer the following research questions:
\begin{enumerate}[leftmargin=0.75cm, itemsep=0.15em]
    \item[Q1.] Can an annotation-free fine-tuning approach yield better performance compared to the unsupervised state-of-the-art approaches? 
    \item[Q2.] How does \texttt{MedUnA} use unpaired images and textual descriptions in contrast to the regular pretraining-finetuning paradigm for medical image classification? 
    \item[Q3.] How does the performance of \texttt{MedUnA} gets affected when the backbone is changed from \texttt{CLIP} to \texttt{MedCLIP}? 
    \item[Q4.]  Are the learned embeddings efficient at the task of multi-class medical image classification?
\end{enumerate}

\vspace{0.1cm}
\noindent\textbf{Datasets:}
Table \ref{tab:datasets_summary} describes the specifics of the datasets: Shenzhen TB (\texttt{S-TB}) \cite{shenzhenTB, jaeger2014two}, Montgomery TB (\texttt{M-TB}) \cite{montgomeryTB, jaeger2014two}, Indian Diabetic Retinopathy 2018 dataset (\texttt{IDRID}) \cite{idrid, porwal2018indian}, the International Skin Imaging Collaboration dataset (\texttt{ISIC}) \cite{isicArchive, codella2018skin} and Guangzhou Pneumonia 
(\texttt{G-Pneumonia}) \cite{guangzhou}.  
% The \texttt{S-TB} dataset comprises chest X-rays from Shenzhen No.3 People's Hospital in China, including normal and tuberculosis-affected lungs. The \texttt{M-TB} dataset, curated by the Department of Health and Human Services, Montgomery, USA, complements \texttt{S-TB} with additional chest X-rays for tuberculosis detection. The \texttt{G-Pneumonia} dataset contains chest X-rays from Guangzhou Women and Children’s Medical Center, China. \texttt{IDRID} includes retinal images with various stages of diabetic retinopathy from an eye hospital in India. The \texttt{ISIC} dataset features a large collection of dermoscopic images of skin lesions, including various skin cancers.

{\renewcommand{\arraystretch}{1.0}
\begin{table}[h!]
\centering
\resizebox{0.5\textwidth}{!}{%
\begin{tabular}{l|ccccc}
    \toprule
    Dataset $\rightarrow$ & \cellcolor[HTML]{FEFEFF}\texttt{S-TB} \cite{shenzhenTB, jaeger2014two} & \cellcolor[HTML]{F4F8F8}\texttt{M-TB} \cite{montgomeryTB, jaeger2014two} & \cellcolor[HTML]{EDF7F8}\texttt{G-Pneumonia}\cite{guangzhou} & \cellcolor[HTML]{E4F8FC}\texttt{IDRID}\cite{idrid, porwal2018indian}  & \cellcolor[HTML]{DBF5FA}\texttt{ISIC}\cite{isicArchive, codella2018skin} \\
    \midrule
    \texttt{\#samples}   & \cellcolor[HTML]{FEFEFF}662 &  \cellcolor[HTML]{F4F8F8}138  & \cellcolor[HTML]{EDF7F8}5,856 & \cellcolor[HTML]{E4F8FC}516 & \cellcolor[HTML]{DBF5FA}11,720\\
    \texttt{\#classes}  & \cellcolor[HTML]{FEFEFF}2 &  \cellcolor[HTML]{F4F8F8}2  & \cellcolor[HTML]{EDF7F8}2 & \cellcolor[HTML]{E4F8FC}5 & \cellcolor[HTML]{DBF5FA}7\\
    \bottomrule
\end{tabular}}
\vspace{-3.5mm}
\caption{\small Summary of number of patients/samples and the number of classes for each dataset.}
\label{tab:datasets_summary}
\vspace{-2mm}
\end{table}}

\begin{figure*}[ht!]
    \centering
    \begin{minipage}[t]{0.32\linewidth}
        \centering
        \includegraphics[height=0.65\textwidth]{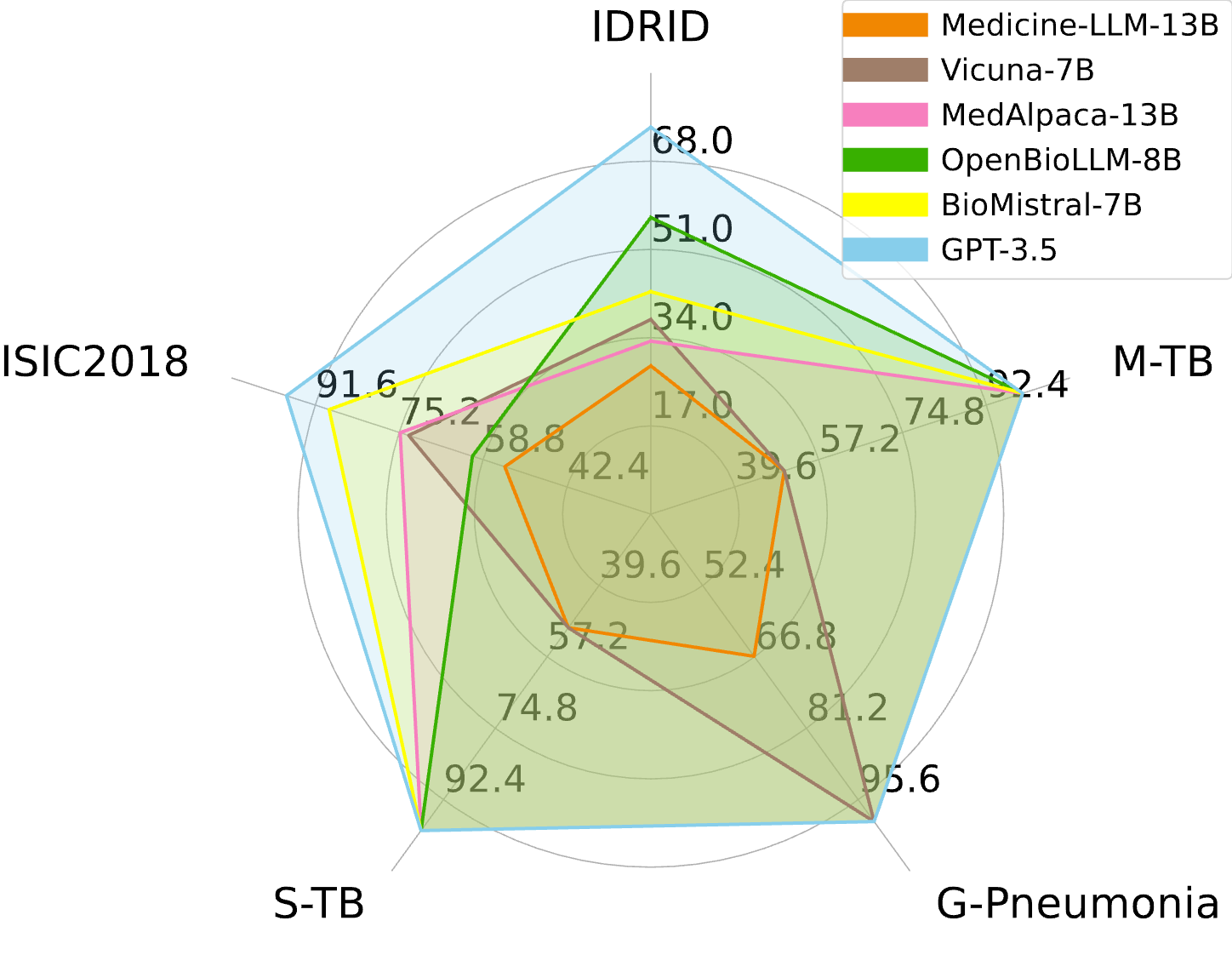}
        \caption{\small Text classifier accuracy across different datasets when descriptions are generated by different \texttt{LLMs}.}
        \label{fig:radar_LLM_text}
    \end{minipage}\hfill
    \begin{minipage}[t]{0.32\linewidth}
        \centering
        \includegraphics[width=\linewidth]{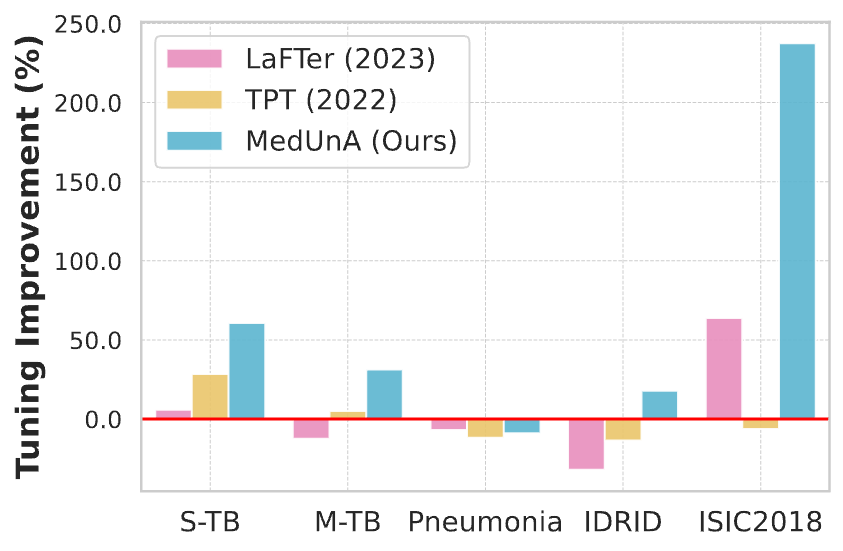}
        \caption{\small Tuning improvement with \texttt{MedCLIP-Swin} Zero-Shot as benchmark. \textcolor{red}{----} denotes the zero-shot CLIP.}
        \label{fig:MedUnAgain}
    \end{minipage}\hfill
    \begin{minipage}[t]{0.32\linewidth}
        \centering
        \includegraphics[width=\linewidth]{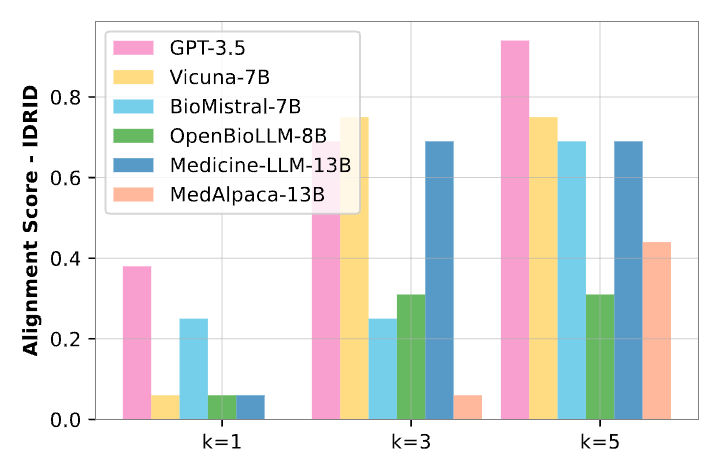}
        \caption{\small Comparison of top-$k$ confident matches between the \textit{unpaired} textual and visual embeddings for IDRID \cite{idrid, porwal2018indian} dataset.}
        \label{fig:k_alignments}
    \end{minipage}
    \vspace{-4mm}
    \label{fig:MedUnA-combined}
\end{figure*}

\vspace{0.1cm}
\noindent\textbf{Implementation:}
We first created a database of textual descriptions tailored to each dataset's pathology classes (e.g., for G-Pneumonia, queries like "Describe a chest x-ray affected by {Pneumonia}"). These unpaired descriptions were encoded into embeddings, then classified by a linear adapter using cross-entropy loss. In the unsupervised regime (Fig. \ref{fig:MedUnA_workflow}), weakly augmented images were used to generate pseudo labels, while strongly augmented ones along with a learnable prompt vector of size (hidden\_size $\times$ num\_tokens) are fed to a visual encoder. The textual classifier then generates logits based on the obtained visual embeddings and we use the training criterion as mentioned in Section \ref{sec:methodology}. Once the learnable prompt vector and the textual classifier are tuned we evaluate the performance on the test set (Q2). The results in Table \ref{tab:MedUnA-acc} compare the performance of \texttt{MedUnA} to other baselines using different visual encoders. We ran experiments using two visual encoders: (1) \texttt{CLIP-ViT-B/32} with a hidden size of 768 and the number of tokens as 50, and (2) \texttt{MedCLIP-Swin} with a hidden size of 96 and the number of tokens as 113. Each dataset was split 60-20-20 for training, validation, and testing. After comparing optimizers, we found that SGD with a learning rate of $10^{-2}$ performed best. We used different LLMs to generate the descriptions in Stage 1. It is evident from Fig. \ref{fig:radar_LLM_text} as well as Table \ref{tab:medUNA_LLM} that GPT-3.5 generated descriptions that enhanced the overall performance of our methodology. The experiments were conducted on 2nd Gen AMD Epyc processors and a single Nvidia A100 Tensor Core GPU.

\vspace{-5mm}
\subsection{Results and Discussion}
\noindent\textbf{MedUnA \textit{vs.} State-of-the-Art:} Table~\ref{tab:MedUnA-acc} reports the Top-1 test accuracy of \texttt{MedUnA}, alongside three carefully selected baselines from distinct fine-tuning groups: (1) zero-shot \texttt{MedCLIP}, (2) unsupervised label-free fine-tuning \texttt{LaFTer}, and (3) test-time adaptation \texttt{TPT}. This selection aims to highlight that \texttt{MedUnA}, which aligns closely with \texttt{LaFTer}, demonstrates superior performance compared to these methods. The reason for comparing \texttt{MedUnA} against diverse baselines is to showcase its strength, contextualize performance, and highlight its advantages. As evident from Fig. \ref{fig:MedUnAgain}, using a \texttt{MedCLIP} backbone gives better results on the medical datasets compared to the \texttt{CLIP} backbone (Q3). Furthermore, \texttt{MedUnA} yields an accuracy gain of 60.7\% compared to the \texttt{MedCLIP} baseline on Shenzhen-TB, 31.25\% on Montgomery-TB, 17.88\% on IDRID and 239\% on \texttt{ISIC} (Q1). Interestingly, we do not observe similar gain on the G-Pneumonia dataset, which can be explained by \texttt{MedCLIP} being trained on the RSNA Pneumonia dataset. Additionally, \texttt{MedUnA} outperforms \texttt{TPT}, showing that adapting prompts at test time alone does not improve generalization.

\noindent\textbf{Correlation between the accuracy of the textual classifier and the overall performance:}
Fig. \ref{fig:radar_LLM_text} plots the accuracy of the textual classifier when trained using the textual descriptions generated by different \texttt{LLMs}. It is evident that the descriptions generated by \texttt{GPT-3.5} lead to a model with a better average performance compared to other \texttt{LLMs}. Since our unsupervised training is dependent on the accuracy of pseudo labels, it is important to use a reliable textual classifier.

\noindent\textbf{Evaluation of LLMs \& Textual-Visual Alignment in MedUnA:}
Our experiments with various LLMs revealed interesting correlations between text classifier accuracy and \texttt{MedUnA} performance (Fig. \ref{fig:radar_LLM_text} and Table \ref{tab:medUNA_LLM}). For example, GPT-3.5 consistently achieved high accuracy and strong performance on datasets like S-TB and M-TB. On G-Pneumonia, GPT-3.5 achieved the highest MedUnA performance (76.12\%), while Medicine-LLM-13B, despite 66.67\% classifier accuracy, had a significantly lower performance (37.5\%). Furthermore, analysis on textual-visual alignment across LLMs shows that medical-specific LLMs generally provide better alignment for higher top-k values compared to general-purpose models like GPT-3.5 and Vicuna. In Fig. \ref{fig:k_alignments} GPT-3.5's peak at k=5 reflects the quality of its generated descriptions.

\begin{figure}[t!]
% \vspace{-2mm}
    \centering
        \includegraphics[height=0.35\linewidth]{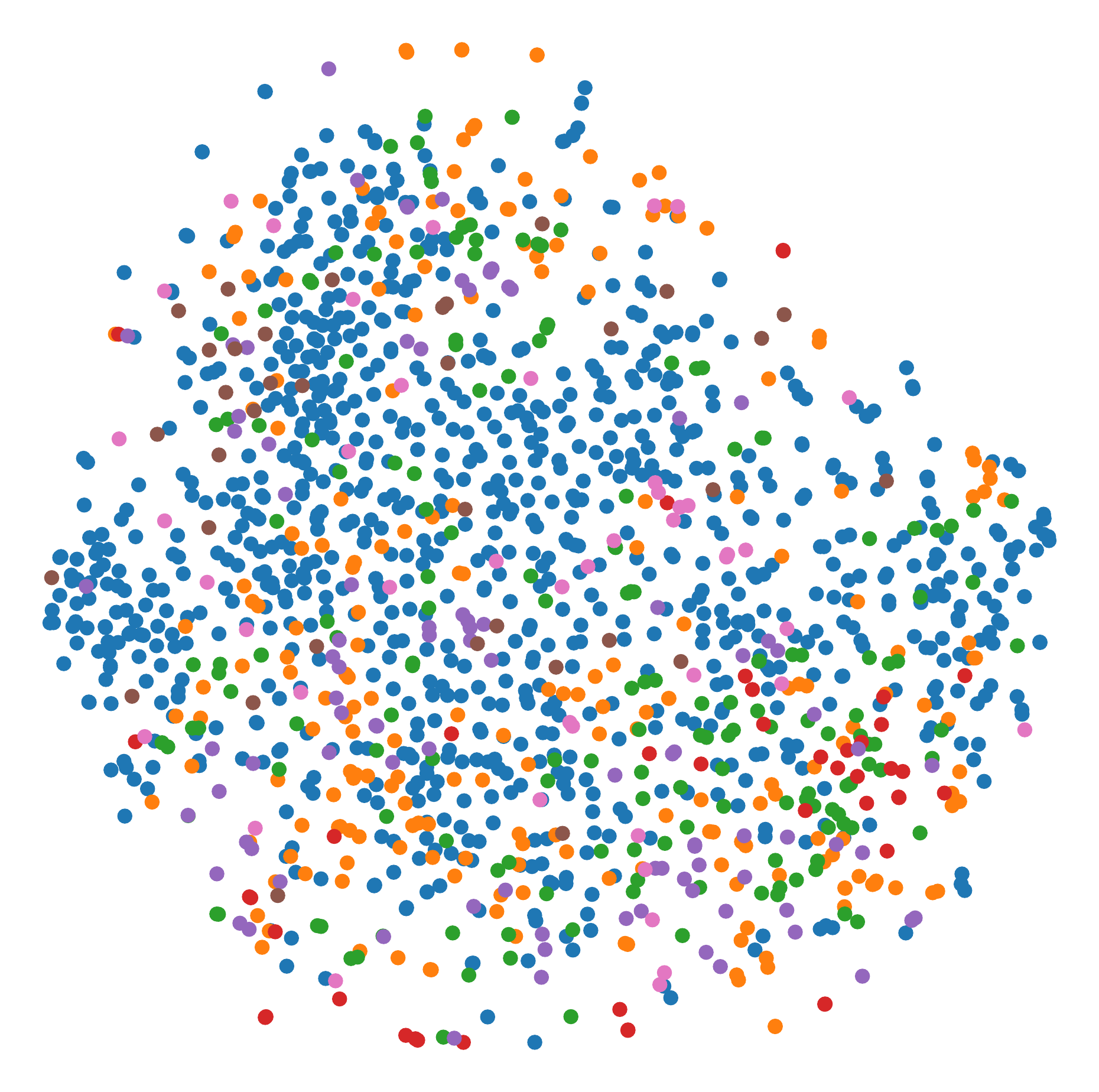}
        \includegraphics[height=0.35\linewidth]{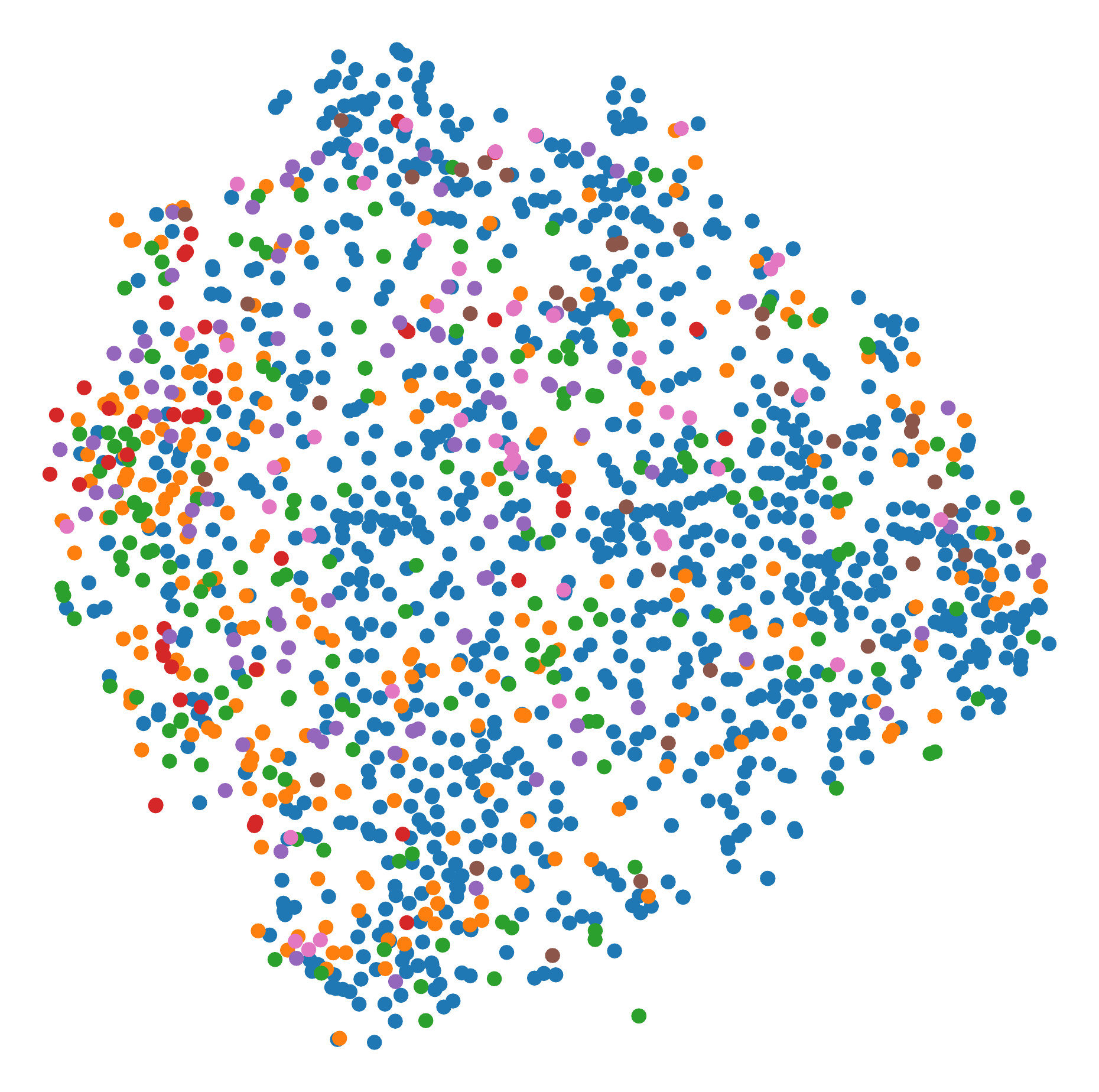}
        \vspace{-2.5mm}
    \caption{\small t-SNE plot of zero-shot \texttt{MedCLIP}  \textit{vs.} our method \texttt{MedUnA} for \texttt{ISIC} dataset having 7 classes.}
    \label{fig:tsne_plot}
    \vspace{-3.5mm}
\end{figure}

\noindent\textbf{Distribution shift of the learned embeddings:}
In Fig.\ref{fig:tsne_plot}, the left plot, evaluating the \texttt{ISIC} dataset with seven classes using zero-shot \texttt{MedCLIP}, shows some overlap between clusters, indicating suboptimal performance for this complex dataset. The right plot demonstrates that \texttt{MedUnA} produces more distinct clusters, suggesting better differentiation (Q4). 

{\renewcommand{\arraystretch}{1.0}
\begin{table}[t!]
\centering
\resizebox{0.5\textwidth}{!}{%
\begin{tabular}{l|ccc}
    \toprule
    Dataset $\downarrow$ & \cellcolor[HTML]{E5F9FC}\texttt{OpenBioLLM-8B} & \cellcolor[HTML]{DEF7FA}\texttt{MedAlpaca-13B} & \cellcolor[HTML]{DBF5FA}\texttt{GPT-3.5} \\
    \midrule
    \texttt{S-TB} \cite{shenzhenTB, jaeger2014two}   & \cellcolor[HTML]{E5F9FC}42.86 &  \cellcolor[HTML]{DEF7FA}42.86  & \cellcolor[HTML]{DBF5FA}\textbf{67.67} \\
    \texttt{M-TB} \cite{montgomeryTB, jaeger2014two} & \cellcolor[HTML]{E5F9FC}48.28 &  \cellcolor[HTML]{DEF7FA}51.72  & \cellcolor[HTML]{DBF5FA}\textbf{72.41} \\
    \texttt{G-Pneumonia} \cite{guangzhou} & \cellcolor[HTML]{E5F9FC}62.50  &  \cellcolor[HTML]{DEF7FA}62.50   & \cellcolor[HTML]{DBF5FA}\textbf{76.12} \\
    \texttt{IDRID} \cite{idrid, porwal2018indian} & \cellcolor[HTML]{E5F9FC}31.07 &  \cellcolor[HTML]{DEF7FA}15.53  & \cellcolor[HTML]{DBF5FA}\textbf{32.04} \\
    \texttt{ISIC} \cite{isicArchive, codella2018skin}  & \cellcolor[HTML]{E5F9FC}08.93  &  \cellcolor[HTML]{DEF7FA}24.10  & \cellcolor[HTML]{DBF5FA}\textbf{29.70} \\
    \midrule
    Average    & \cellcolor[HTML]{E5F9FC}38.29 & \cellcolor[HTML]{DEF7FA}45.34 & \cellcolor[HTML]{DBF5FA}\textbf{55.79} \\
    \bottomrule
\end{tabular}}
\vspace{-3.5mm}
\caption{\texttt{MedUnA} performance ( accuracy \%) when using the descriptions generated by different open-source LLM models.}
\label{tab:medUNA_LLM}
\vspace{-4mm}
\end{table}}

% \begin{figure}[t!]
% % \vspace{-2mm}
%     \centering
%         \includegraphics[height=0.37\linewidth]{figs/tSNE_zero_shot_ISIC2018.pdf}
%         \includegraphics[height=0.37\linewidth]{figs/tSNE_MedUnA_ISIC2018.pdf}
%         \vspace{-2.5mm}
%     \caption{\small t-SNE plot of zero-shot \texttt{MedCLIP} \textit{vs.} our method \texttt{MedUnA} \texttt{ISIC} dataset having 7 classes.}
%     \label{fig:tsne_plot}
%     \vspace{-3.5mm}
% \end{figure}

\vspace{-4mm}
\section{Conclusion}
\vspace{-3mm}
To address the challenge of extensive pre-training requirements of VLMs on vast medical datasets and to overcome the constraint of using paired image-text data, our two-stage language-guided approach \texttt{MedUnA} effectively integrates unpaired and unlabeled visual and textual descriptions to reduce the modality gap to enhance the overall performance. It makes optimal use of the more readily available and less restricted textual data than paired labeled images. \texttt{MedUnA} is a pioneering attempt in the direction of minimizing the need for extensive paired and labeled data and developing a method that is more scalable and applicable to novel disease classes. 

\section{Compliance with Ethical Standards}
This research study was conducted retrospectively using human subject data made available in open access by \cite{shenzhenTB}, \cite{jaeger2014two}, \cite{montgomeryTB}, \cite{guangzhou}, \cite{idrid}, \cite{porwal2018indian}, \cite{isicArchive}, \cite{codella2018skin}. Ethical approval was not required as confirmed by the license attached with the open access data.

\bibliographystyle{ieeetr}
\bibliography{main}

\end{document}